\documentclass{article}

\usepackage{PRIMEarxiv}

\usepackage[utf8]{inputenc}    
\usepackage[T1]{fontenc}       
\usepackage{hyperref}          
\usepackage{url}               
\usepackage{booktabs}          
\usepackage{amsfonts}          
\usepackage{nicefrac}          
\usepackage{microtype}         
\usepackage{graphicx}          
\usepackage{tabularx}          
\usepackage{fancyhdr}          
\usepackage{lipsum}            
\usepackage{amsmath}
\usepackage{multirow}
\usepackage{multicol}

\usepackage{tikz}
\usepackage{pifont}
\usepackage[table]{xcolor} 
\usepackage{pifont}        
\usepackage{array}

\graphicspath{{media/}}     

\title{CompressNAS : A Fast and Efficient Technique for Model Compression using Decomposition} 

\author{
  {\bf Sudhakar Sah, Nikhil Chabbra, Matthieu Durnerin } \\ \\
  STMicroelectronics\\
  Toronto, Canada \\
  \texttt{sudhakar.sah@st.com}
}
\begin{document}

\maketitle

\begin{abstract}
Deep Convolutional Neural Networks (CNNs) are increasingly difficult to deploy on microcontrollers (MCUs) and lightweight NPUs (Neural Processing Units) due to their growing size and compute demands. Low-rank tensor decomposition, such as Tucker factorization, is a promising way to reduce parameters and operations with reasonable accuracy loss. However, existing approaches select ranks locally and often ignore global trade-offs between compression and accuracy. We introduce \emph{CompressNAS}, a MicroNAS-inspired framework that treats rank selection as a global search problem. \emph{CompressNAS} employs a fast accuracy estimator to evaluate candidate decompositions, enabling efficient yet exhaustive rank exploration under memory and accuracy constraints. In ImageNet, \emph{CompressNAS} compresses ResNet-18 by $8\times$ with less than $4\%$ accuracy drop, in COCO, we achieve $2\times$ compression of YOLOv5s without any accuracy drop and $2\times$ compression of YOLOv5n with $2.5\%$ drop. Finally we present a new family of compressed models \emph{STResNet} with competitive performance compared to other efficient models.

  \keywords{TinyML \and lightweight CNNs \and EdgeAI \and Model Compression }
\end{abstract}

\section{Introduction}
\label{sec:intro}

Deep convolutional neural networks (CNNs) have become the backbone of modern computer vision, powering applications ranging from classification and detection to segmentation. Yet, these accuracy gains are obtained at the expense of rapidly increasing model depth, parameter count, and computational demand. This poses a major obstacle for deployment on resource-constrained platforms such as microcontrollers (MCUs) and lightweight neural processing units (NPUs), where limited memory, strict energy budgets, and real-time constraints dominate \cite{sze2020efficient,warden2019tinyml}. Addressing this mismatch between model complexity and hardware constraints has spurred extensive research in neural network compression.

Classical compression methods include pruning redundant weights or filters \cite{han2015deep,li2017pruning}, quantizing weights and activations to low precision \cite{hubara2017quantized,jacob2018quantization}, and distilling knowledge from large teacher models into smaller students \cite{hinton2015distilling}. Although effective, these approaches often require costly retraining and may not guarantee stable accuracy under strict deployment budgets. An alternative line of work leverages low-rank tensor decomposition to directly factorize convolutional kernels, thereby reducing multiply–accumulate (MAC) operations and memory footprint. Pioneering methods based on  Canonical polyadic (CP), Tucker, and tensor-train (TT) decompositions have demonstrated promising results for compressing convolution layers with minimal accuracy loss \cite{lebedev2014speeding,denton2014exploiting,kim2016compression,novikov2015tensorizing}. Among these, Tucker decomposition has emerged to be particularly attractive due to its ability to control ranks across different tensor modes, balancing compression and expressiveness \cite{kolda2009tensor}.

A central challenge in decomposition-based compression is selecting appropriate ranks for each layer. Current pipelines often rely on analytic estimators such as Empirical Variational Bayes Matrix Factorization (EVBMF) \cite{nakajima2013global} or heuristics based on local reconstruction error \cite{lebedev2014speeding,kim2016compression}. While lightweight, these estimators optimize local approximation fidelity rather than global task performance, and they do not naturally account for inter-layer dependencies or end-to-end constraints such as latency or memory budgets. This gap motivates search-based strategies that can identify globally consistent rank configurations.

Recent advances in neural architecture search (NAS) suggests to treat rank selection as an architectural design problem. NAS methods have shown success in identifying compact architectures under hardware-aware objectives \cite{elsken2019neural,cai2020once}. In parallel, zero-cost (ZC) proxies have been developed to estimate network accuracy without training by using gradient- or saliency-based measures from a single forward or backward pass \cite{mellor2021neural,abdelfattah2021zero,tanaka2020pruning}. These proxies correlate strongly with final trained accuracy, and recent work has explored ensembling and symbolic-regression approaches to further improve their predictive power \cite{aznas2024cvpr,ensembleZC2025,craftingZC2024}. Despite this progress, their application to rank selection in tensor decompositions remains underexplored. 

This paper proposes \emph{CompressNAS}, a microNAS framework for Tucker decomposition–based CNN compression. \emph{CompressNAS} formulates rank selection as a global search problem. Our contributions are:
\begin{itemize}
    \item We introduce a lightweight estimator to approximate the decomposition impact on accuracy and compression ratio enabling fast evaluation without finetuning.
    \item We develop an efficient exhaustive search technique that identifies globally consistent rank configurations under model size and accuracy constraints, suitable for MCU/NPU deployment.
    \item We propose an Integer Linear Programming (ILP) search based NAS given the hardware budget. 
    \item We present a new family of compressed backbone \emph{STResNet} having competitive performance in $<4$M Parameter range.   
\end{itemize}

To the best of our knowledge, prior work on decomposition has not reported extreme compression (order of 10$\times$) for classification models such as ResNet, nor any promising results for detection models such as YOLOv5.

\section{Related Work}
\label{sec:related_work}

\paragraph{Tensor Decomposition for CNN Compression.}
Recent work has renewed interest in tensor factorizations for compact CNNs. Beyond classical CP/TT approaches, Tucker and its variants have seen steady progress. Knowledge-Based Systems (2022) introduced Tucker with nonlinear response modeling for accuracy-aware compression \cite{gabor2022kbs_tucker}. Hierarchical Tucker-2 (HT-2) further reduces storage by recursively factorizing the core, improving efficiency on ImageNet-scale models \cite{zdunek2023ht2}. Concurrently, alternating optimization focussing under Tucker was proposed to combine fine-tuning with decomposition steps \cite{liu2025accuracypreserving}. Complementary efforts studied hardware-aware acceleration pipelines for Tucker-compressed CNNs \cite{xiang2023tdc} and training-in-the-low-rank space to stabilize optimization \cite{li2024elrt}.

\paragraph{Automatic and Budget-Aware Rank Selection.}
Selecting Tucker ranks remains central to the accuracy--efficiency trade-off. Early heuristics (e.g., VBMF and sensitivity) have been superseded by learning- or search-based strategies. BATUDE (AAAI 2022) integrates \emph{automatic} rank selection into training under explicit budget constraints, yielding globally consistent rank configurations across layers \cite{batude2022aaai}. More recently, unified frameworks cast rank selection as a continuous search with composite compression losses, enabling training-free or data-light rank discovery \cite{unified2024arxiv}. Layer-interaction–aware schemes have also appeared, arguing that ranks should be chosen with cross-layer coupling in mind \cite{kokhazadeh2025cnn}.

\paragraph{Zero-Cost Proxies and NAS-Inspired Rank Search.}
ZC proxies evaluate architectures without training and using just one minibatch or even a single forward/backward pass, e.g., SynFlow, SNIP, GraSP, and their learned/ensembled variants \cite{abdelfattah2021zero,aznas2024cvpr,ensembleZC2025}. These proxies have matured with aggregation and symbolic-regression–based design to improve model rank correlation with final accuracy \cite{aznas2024cvpr,craftingZC2024}. Despite this progress, their application to \emph{tensor-rank selection} for Tucker remains underexplored: most Tucker works still rely on EBVMF \cite{nakajima2013global}, local reconstruction-error heuristics, per-layer sensitivities, or budget-constrained training loops \cite{batude2022aaai,unified2024arxiv} without considering the global impact on parameter reduction and accuracy trade-off. Our work bridges these lines by framing Tucker rank selection as a NAS problem and leveraging a custom (but simple) ZC proxy to guide ranks across modes and layers, avoiding expensive inner-loop fine-tuning while preserving global budget feasibility.

\paragraph{Efficient models.}
Lightweight CNNs such as MobileNet \cite{howard2017mobilenets}, SqueezeNet \cite{iandola2016squeezenet}, ShuffleNet \cite{zhang2018shufflenet}, and EfficientNet \cite{tan2019efficientnet} have been widely adopted for deployment in resource-constrained environments. These models achieve competitive accuracy with significantly fewer parameters and reduced computational cost compared to conventional architectures. MobileNets employ depthwise separable convolutions to lower FLOPs, SqueezeNet introduces fire modules for parameter efficiency, ShuffleNet leverages channel shuffling with grouped convolutions to enhance information flow, and EfficientNet uses compound scaling to balance network depth, width, and resolution. Together, these architectures form the basis of much of the current research in efficient model design and compression.

\paragraph{Positioning.}

BATUDE \cite{batude2022aaai} leverages Tucker decomposition for compressing CNN models while providing budget-aware optimization. It performs layer-wise rank selection using heuristics and is suitable for memory-constrained deployment. However, the rank selection is fixed once computed and does not support flexible reuse for multiple compression levels.

Accuracy-Preserving Neural Network Compression via Tucker Decomposition \cite{kim2016tucker}  focuses on solving a training-decomposition subproblem iteratively but lacks a global search mechanism and configurable reuse across multiple compression levels.

Unified Framework for Neural Network Compression via Decomposition and Optimal Rank Selection \cite{aghababaeiharandi2024unified} proposes a fine-grain search strategy to select optimal ranks globally across the network. While effective in exploring the compression–accuracy trade-off, it requires repeated optimization for different compression targets, which increases computational overhead.

\emph{CompressNAS} combines the benefits of prior methods by performing global rank search, supporting budget-aware and fine-grain optimization, and introducing a configurable optimization framework: the search is executed once and reused to generate models at different compression levels. Unlike prior approaches, \emph{CompressNAS} reduces optimization overhead, provides flexibility in compression targets, and ensures practical deployment in resource-constrained scenarios.

\section{Tucker Decomposition}
\label{sec:methodology}

\begin{figure}
    \centering
    \includegraphics[width=0.5\linewidth]{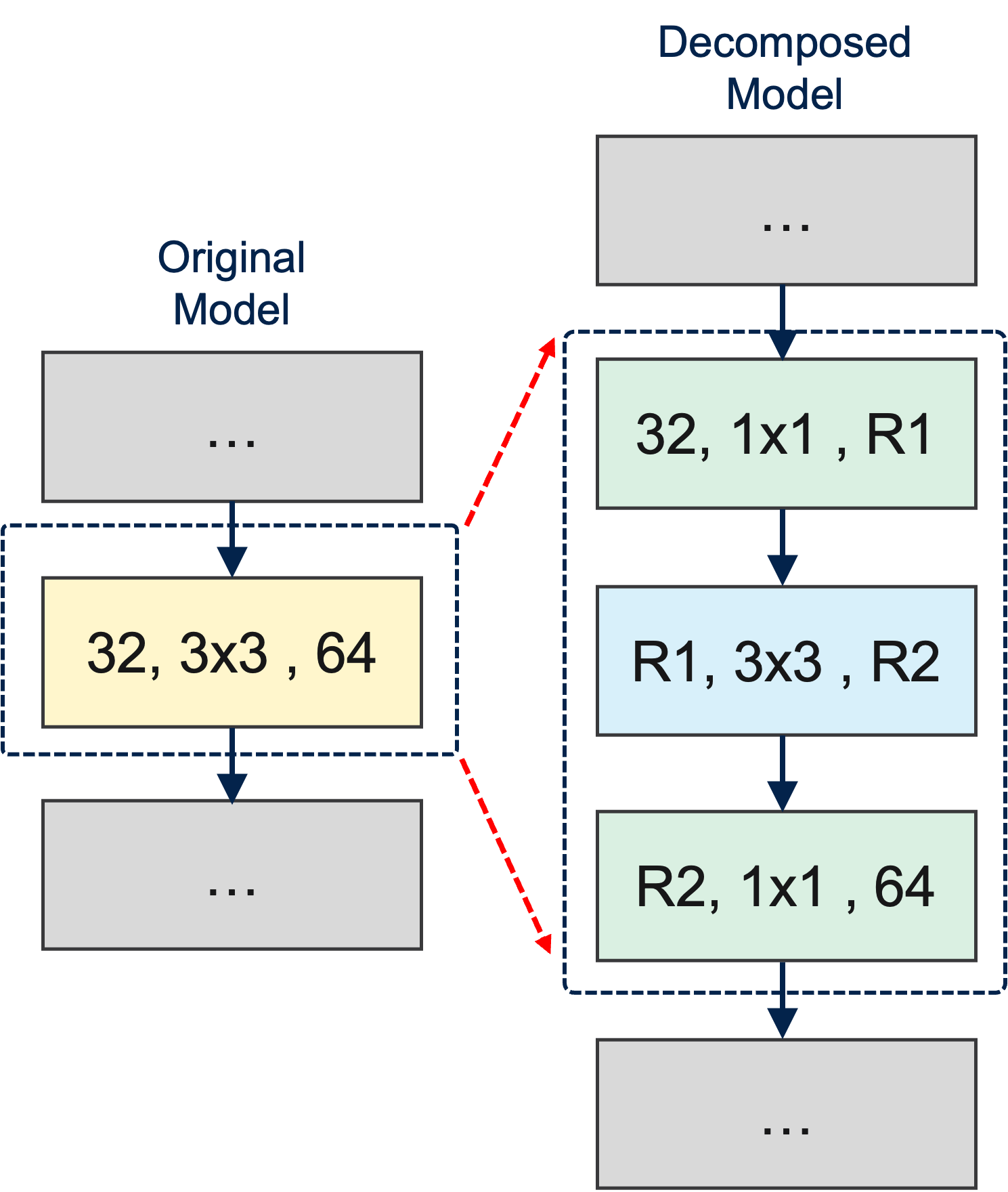}
    \caption{Decomposition of a CNN layer : $R1$ and $R2$ are selected ranks for decomposed layers}
    \label{fig:cnn_decomposition}
\end{figure}

Given a 4D convolutional kernel $\mathcal{W} \in \mathbb{R}^{O \times I \times H \times W}$, where $O$ and $I$ denote the number of output and input channels respectively, and $H \times W$ represents the spatial kernel size, Tucker decomposition factorizes $\mathcal{W}$ into a smaller core tensor $\mathcal{G}$ and projection matrices along each mode:
\begin{equation}
    \mathcal{W} \approx \mathcal{G} \times_1 U^{(1)} \times_2 U^{(2)} \times_3 U^{(3)} \times_4 U^{(4)},
\end{equation}
where $\times_n$ denotes the $n$-mode product and $U^{(n)}$ are factor matrices.  
This decomposition reduces redundancy in the kernel representation and allows replacing a single high-dimensional convolution with a sequence of lower-rank convolutions, thereby reducing both parameter count and floating-point operations (FLOPs).

When applied to CNN layers, Tucker decomposition is typically performed along the input and output channel dimensions while keeping the spatial kernel sizes intact. This strategy enables significant compression without severely impacting accuracy. Empirical studies have shown that Tucker-based factorization of convolutional layers leads to substantial acceleration while maintaining competitive performance \cite{kim2016tucker}.

Figure \ref{fig:cnn_decomposition} shows an example of the decomposition of a layer from CNN model. A layer with $32$ input channels and $64$ output channels is decomposed into three layers. 1. A layer with kernel size 1$\times$1 and $32$ input channels and $R1$ output channels. 2. Second layer (also called core layer) with input channel as $R1$ and output channels as $R2$ where $R1$ and $R2$ are selected input and output ranks. 3. Final layer with input channels as $R2$ and output channel as $64$. The reduction is number of parameters is calculated using Equation \ref{eq:parameter_saving_decomp}. Using same input and output rank as 8 by replacing $R1$ and $R2$ by 8 in Equation \ref{eq:parameter_saving_decomp} can save 17,088 parameters.

\begin{equation}
\label{eq:parameter_saving_decomp}
\text{Number of Parameters} = 32 \cdot 3\times3 \cdot 64 - \big(32 \cdot 1\times1 \cdot R_1 + R_1 \cdot 3\times3 \cdot R_2 + R_2 \cdot 1\times1 \cdot 64 \big)
\end{equation}

Tucker decomposition leverages EVBMF \cite{kim2016tucker} to automatically determine the ranks for each layer automatically. By estimating optimal (but local) ranks for each layer, number of parameters and FLOPS can be reduced significantly while preserving accuracy. However, EVBMF performs layer-wise, independent rank estimation and does not consider global trade-offs across layers, which can lead to suboptimal accuracy–compression balance compared to methods that perform network-wide rank search. Despite this limitation, EVBMF-based Tucker decomposition remains a widely used baseline for low-rank compression due to its simplicity and automatic rank selection. However, model compression using Tucker decomposition is still an open global rank optimization problem with accuracy/compression tradeoff.

\section{CompressNAS Architecture}
\label{sec:quark}

\subsection{Network Proposals}
\begin{figure*}
    \centering
    \includegraphics[width=\textwidth]{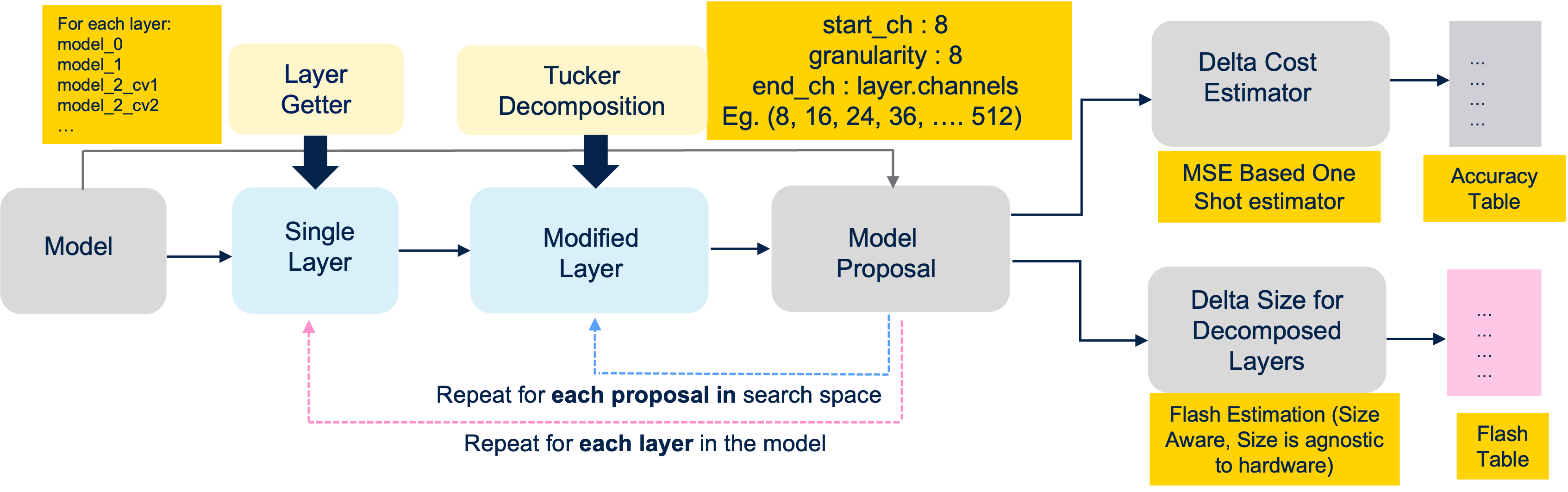}
    \caption{\emph{CompressNAS} : Model Proposal Generation and Profiling}
    \label{fig:quark_proposal_generation}
\end{figure*}

Figure \ref{fig:quark_proposal_generation} shows the complete architecture of {\emph{CompressNAS}. {\emph{CompressNAS} adopts a microNAS-inspired approach in which impact of optimization of each layer is considered independently. 
For convolutional layers, the number of decomposition proposals is determined by the number of output channels. We use an exhaustive search strategy to find the best rank by generating rank proposals beginning from a configurable value (default: 8 channels) and increment in configurable steps of 8 
(e.g., 8, 16, \ldots, up to the maximum number of output channels) or start with 4 channels and increment at the internal of 2,4,8 etc. Lower granularity of filter means larger search space and theoretically fine grain optimized model.   

For each rank proposal, Tucker decomposition \cite{kim2016tucker} is applied to the target layer, 
and the original layer is replaced by decomposed layer as shown in figure~\ref{fig:cnn_decomposition}. 
Each substitution yields a model candidate, i.e., one decomposed layer with one rank proposal corresponds 
to one model proposal. For every candidate model, we estimate its impact on accuracy (\(\Delta acc\)) 
and memory footprint or flash size (\(\Delta flash\)). 

This process is repeated across all the convolutional and linear layers. 
For instance, a network with $20$ candidate layers and 10 rank proposals per layer creates $200$ candidate models. Since the number of channels for each layer is not constant, proposals per layer is variable.  
The output of this procedure is summarized in two lookup tables: 
one for \(\Delta acc\) and another for \(\Delta flash\) 

\subsection{Accuracy Estimator}
Accurately estimating the performance of a model after layer decomposition is critical to evaluating the effectiveness of candidate architectures. We investigated a variety of existing zero-cost proxies, including NASWOT \cite{mellor2021neural}, GraSP, SNIP, and ZiCo \cite{abdelfattah2021zero}, thereby covering both activation-based and gradient-based approaches. Figure~\ref{fig:zc_score_comparison} illustrates the difference between the proxy scores of the reference model and its decomposed variants across different ranks (x-axis) for the first seven layers of the YOLOv5n model. The leftmost point in each plot corresponds to the reference model and, as expected, yields a zero difference. Subsequent points represent the score differences of candidate models after decomposition at increasing ranks.Ideally, higher ranks should correspond to higher zero-cost proxy scores, reflecting improved accuracy of the decomposed model. However, the graph indicates large inconsistencies, failing to capture the expected monotonic trend making these proxies unreliable for estimate the model accuracy of decomposed models. 

As an alternative, we employed a mean squared error (MSE)-based proxy, computed by treating the feature vector output of the decomposed layer as the prediction and the feature vector output of the corresponding reference layer as the target, as illustrated in Figure~\ref{fig:mse_proxy}. Unlike existing zero-cost proxies, the MSE-based estimator consistently exhibits the expected behavior, with higher scores observed at higher ranks as shown in figure \ref{fig:zc_score_comparison}. This suggests that the proposed proxy provides a more reliable measure of the impact on accuracy of decomposition compared to existing methods.

\begin{figure*}
    \centering
    \includegraphics[width=\textwidth]{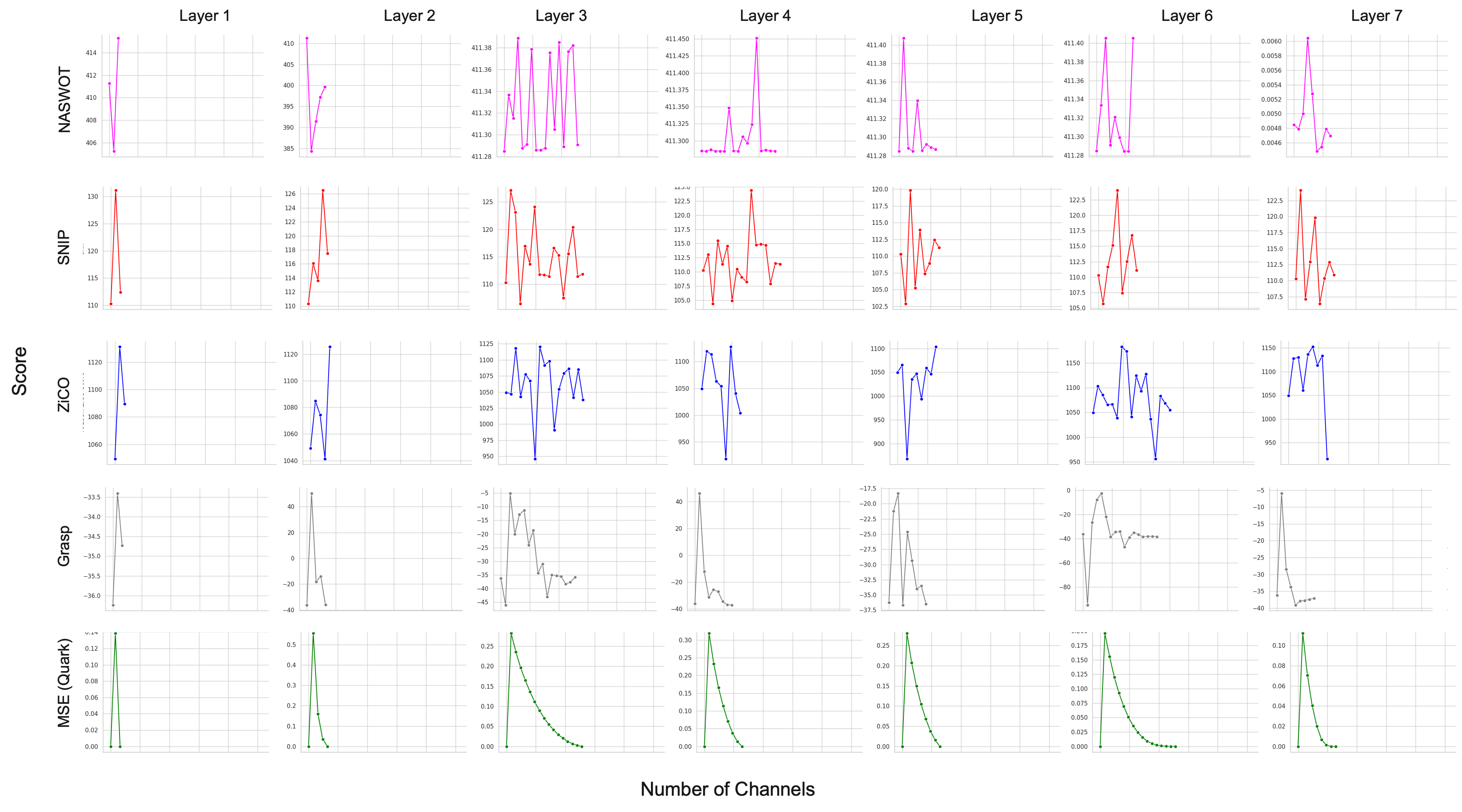}
    \caption{Comparison of different zero cost estimator scores for 7 initial layers of YOLOv5n}
    \label{fig:zc_score_comparison}
\end{figure*}

\begin{figure}
    \centering
    \includegraphics[width=0.6\linewidth]{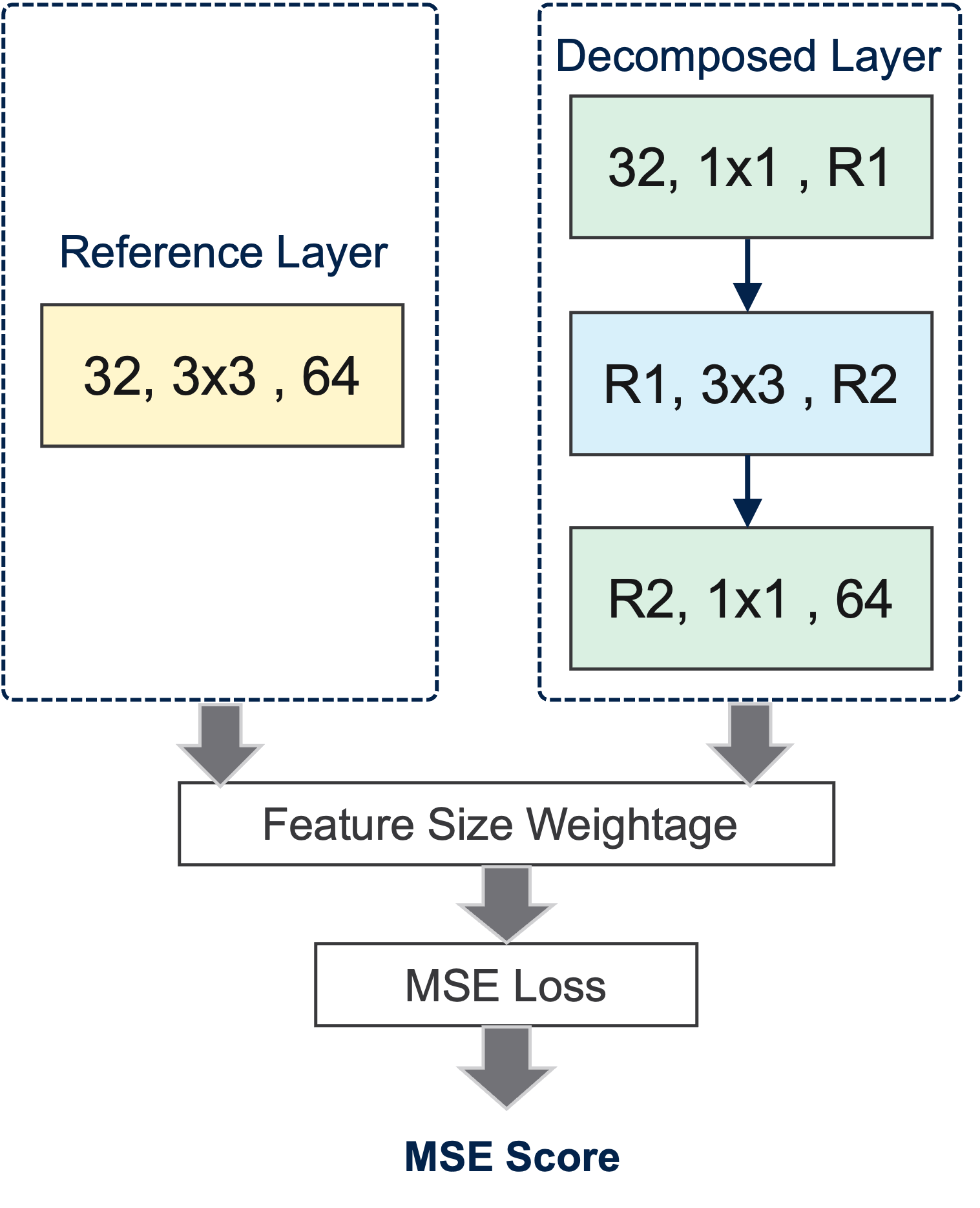}
    \caption{Mean Square Error based Accuracy Estimator}
    \label{fig:mse_proxy}
\end{figure}

\subsection{Flash Estimator}
After every layer replacement, the model is exported to ONNX \cite{onnx2019} and the difference between the reference model and the candidate model provides $\Delta flash$ for a candidate model. We also explored a much simpler hardware agnostic approach to make this process extremely fast by using theoretical calculation of number of parameters for candidate layer and reference layer. For eg. a layer with $M$ output channels, $N$ input channel, $k$ as kernel size and $R$ as decomposed rank, the delta flash can be calculated using equation\ref{eq:delta_flash}. We can use either approach but in case of MCUs, hardware aware method is more reliable.

\begin{equation}
\label{eq:delta_flash}
\Delta flash = N M k^2 - \left(NR\cdot 1\!\times\!1 + R^2 \cdot 3\!\times\!3 + RM\cdot 1\!\times\!1 \right)
\end{equation}

\subsection{Neural Architecture Search}
Using the accuracy estimator, the flash estimator, and MicroNAS strategy, we build two tables: \textbf{accuracy table} and \textbf{flash table}. In each table, rows represent convolution layers, columns represent rank values, and each cell coressponds to the change ($\Delta$) in accuracy or flash compared to the reference model for that rank. We then formulate the problem of searching for compressed model as an Integer Linear Programming (ILP) \cite{trauth1969integerlinear} optimization problem. The objective is to maximize accuracy and while adhering to flash constraints as equation \ref{eq:ilp_formulation}. In the ILP formulation, one variable vector represents the indices of the layers where decompositions are applied, and the other represents the candidate (rank for decomposition) for those layers. We used the open-source PuLP library \cite{PuLP} to solve this ILP problem. The result is a ranking list of the top \textit{k} model proposals. ILP search provides top model candidates satisfying the flash and accuracy budget. The result is stored as proposed rank for each layer.

\begin{equation}
\label{eq:ilp_formulation}
\begin{aligned}
Accuracy = \max \quad & \sum_{(i,j) \in E} \Delta \text{accuracy}_{ij} \\
\text{s.t.} \quad & \sum_{(i,j) \in E} \Delta \text{flash}_{ij} \leq flash_{\max}.
\end{aligned}
\end{equation}

\subsection{Optimized Model Architecture}
This is the final step of model optimization and as depicted in Figure \ref{fig:quark_optimized_model_creation}, \emph{CompressNAS} uses the reference model and layer wise rank proposal to iteratively replace each layer with decomposed layer resulting in \emph{CompressNAS} Optimized Model. This model needs full retraining to achieve highest possible accuracy. In case a new model is needed with different flash and/or accuracy constraint, the optimization process does not need to be repeated, instead, only ILP search is performed which takes few seconds to generate a new proposal but this model again needs full retraining.

\begin{figure*}
    \centering
    \includegraphics[width=\textwidth]{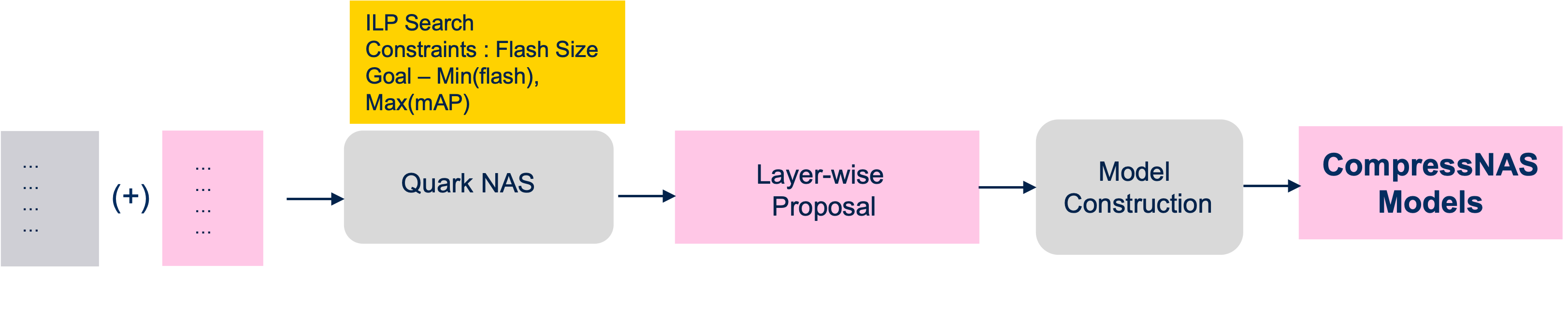}
    \caption{\emph{CompressNAS} : Optimized Model Creation}
    \label{fig:quark_optimized_model_creation}
\end{figure*}

\section{Results}
\label{sec:results}

We compared performance of classification models on ImageNet \cite{deng2009imagenet} and Flowers102 \cite{nilsback2008automated_flowers102} datasets using default timm \cite{rw2019timm} training pipeline. Object detection models are trained using ultralytics \cite{ultralytics2023yolo} library on MS COCO \cite{lin2014coco} datasets.

\subsection{Optimized Models}
\begin{table}[h]
\centering
\caption{Performance of Classification models trained on Imagenet datasets under different compression budgets. ({\textsuperscript{*}}with NAS, {\textsuperscript{+}}without NAS, NAS uses finetuning for each layer and it is very slow)}
\label{tab:resnet18_results}
\begin{tabular}{l@{\hskip 2pt}c@{\hskip 2pt}c@{\hskip 2pt}c}
\hline
\textbf{Model} & \textbf{Params (M)} & \textbf{Top-1 (\%)} & \textbf{Compression} \\ \hline
ResNet18 & 11.68 & 70.50& 1.00x \\
Tucker{\textsuperscript{*}} & 1.66  & 66.62 & $7.03\times$ \\
Tucker{\textsuperscript{+}} & 1.15  & 58.79 & $10.15\times$ \\ 
\textbf{CompressNAS} & 1.50  & 66.07 & $7.78\times$ \\ \hline
ResNet34 & 21.79 & 76.41 & 1.00$\times$ \\
Tucker{\textsuperscript{+}} & 1.01  & 57.35 & $21.57\times$ \\
\textbf{CompressNAS} & 1.08  & 59.42 & $20.18\times$ \\ \hline
MobilenetV2 & 3.50 & 71.36 & $1.00\times$  \\
Tucker{\textsuperscript{*}} & 1.56 & 26.45 & $2.24\times$ \\ 
\textbf{CompressNAS} & 1.57 & 38.58 & $2.24\times$  \\ \hline

\end{tabular}
\end{table}

\begin{table}[h]
\centering
\caption{Performance of YOLOv5 models trained on COCO dataset under different compression budgets. ({\textsuperscript{*}}without NAS)}
\label{tab:yolov5n_results}
\begin{tabular}{lccc}
\hline
\textbf{Model} & \textbf{Params (M)} & \textbf{mAP} & \textbf{Compression} \\ \hline
YOLO5n   & 1.95 & 24.20 & $1.00\times$ \\
Tucker{\textsuperscript{*}} & 0.95 & 16.60 & $2.03\times$ \\
\textbf{CompressNAS}   & 0.96 & 21.70 & $2.03\times$ \\
 \hline
YOLO5s & 7.03 & 32.99 & $1.00\times$ \\
Tucker{\textsuperscript{*}} & 2.99 & 25.90 & $2.35\times$ \\
\textbf{CompressNAS} & 3.49 & 32.90 & $2.00\times$ \\
\textbf{CompressNAS} & 2.11  & 25.80 & $3.33\times$ \\ \hline
\end{tabular}
\end{table}

\begin{table}[h!]
\centering
\caption{Extreme compression results on ResNet18 with smaller dataset (Flowers102 \cite{nilsback2008automated_flowers102})}
\label{tab:resnet18_flowers102_results}
\begin{tabular}{l@{\hskip 2pt}c@{\hskip 2pt}c@{\hskip 2pt}c}
\hline
\textbf{Model} & \textbf{Params (M)} & \textbf{Top-1 (\%)} & \textbf{Compression} \\
\hline
Reference & 10.73 & 91.96 & 1.00$\times$ \\
CompressNAS & 5.36 & 92.06 & 2.00$\times$ \\
CompressNAS & 3.21 & 91.54 & 3.33$\times$ \\
CompressNAS & 1.07 & 89.32 & 10.00$\times$ \\
CompressNAS & 0.85 & 88.43 & 12.5$\times$ \\
\hline
\end{tabular}

\end{table}

Tables~\ref{tab:resnet18_results}--\ref{tab:resnet18_flowers102_results} compare \emph{CompressNAS} with vanila Tucker decomposition \cite{kim2016tucker} ~\cite{kim2016tucker,lebedev2014speeding} and Tucker with Neural Architecture Search (NAS) \cite{Sankaran2021Neutrino} under varying compression budgets. The NAS method uses search algorithm to decide whether to decompose a particular layer or not by estimating the impact on accuracy by the means of finetuning for few epochs. The NAS also helps to find the best configuration for a given flash constraint.  

As show in Table~\ref{tab:resnet18_results}, on \textbf{ResNet-18 \cite{he2016deep} (ImageNet)}, tucker with NAS achieves a $7.03\times$ compression ratio but suffers a 4.9-point Top-1 accuracy drop, while tucker without NAS collapses to only 58.8\% Top-1 accuracy at $7.5\times$ compression. In contrast, \emph{CompressNAS} sustains 66.1\% Top-1 accuracy at a comparable $7.78\times$ compression, showing better robustness under aggressive reduction.  

As show in Table~\ref{tab:resnet18_results}, for \textbf{ResNet-34} \cite{he2016deep}, with NAS compresses the model to $21.6\times$ but degrades Top-1 accuracy to 57.4\%. \emph{CompressNAS} achieves a similar $20.2\times$ compression but retains a higher Top-1 accuracy of 59.4\%, again demonstrating improved accuracy preservation.  

As show in Table~\ref{tab:resnet18_results}, for \textbf{MobileNetV2} \cite{sandler2018mobilenetv2}, Tucker severely underperforms (26.5\% Top-1 at $2.24\times$ compression). \emph{CompressNAS} recovers performance, reaching 38.6\% Top-1 accuracy at the same compression budget, however, absolute top-1 accuracy of optimized MobilenetV2 model is very low and this highlights the fact that these compression algorithms fail on hand crafted smaller models like MobileNetV2. 

Table \ref{tab:yolov5n_results} shows results for \textbf{YOLOv5n} \cite{ultralytics2023yolo} trained on COCO \cite{lin2014coco} dataset, tucker based compression results in sharp mAP drops (e.g., from 24.2\% to 16.6\%). \emph{CompressNAS} improves accuracy at identical compression (21.7\% mAP at $2.03\times$ for \textbf{YOLOv5n}, 32.9\% at $2\times$ for \textbf{YOLOv5s}) and remains competitive even under more aggressive reduction ($3.33\times$). It is important to note that YOLO5s got compressed by $2\times$ without any accuracy drop.  

Finally, Table~\ref{tab:resnet18_flowers102_results}  shows \textbf{extreme compression results on Flowers102 \cite{nilsback2008automated_flowers102} with ResNet-18} , \emph{CompressNAS} maintains accuracy under $3.33\times$ compression (91.5\% vs. 91.9\% baseline) and only gradually degrades under extreme $10\times$ and $12.5\times$ compression, where Top-1 drops to 89.3\% and 88.4\%, respectively.  

\begin{table*}[h!]
\centering
\caption{Comparison of state-of-the-art compression methods on ResNet-18.}
\label{tab:quark_sota_comparison}

\rowcolors{2}{gray!10}{white}  

\begin{tabular}{l|ccc|p{1.5cm}p{2cm}p{2.2cm}}
\hline
\rowcolor{blue!10}
\textbf{Technique} & \textbf{Top-1(\%)}  & \textbf{Top-5(\%)}  & \textbf{Compression} & \textbf{Budget-aware} & \textbf{Fine-grained ranks} & \textbf{Configurable Optimization} \\
\hline
Torchvision\cite{torchvision2016,he2016deep} & 69.75 & 89.08 & 1.00$\times$ &  &  &  \\
Vanilla Tucker (2016) \cite{kim2016tucker} & - & 87.53 & 2.25$\times$ & \ding{55} & \ding{55} & \ding{55} \\
MUSCO (2019) \cite{gusak2019musco} & - & 88.78 & 2.42$\times$ & \ding{55} & \ding{55} & \ding{55} \\
Stable EPC (2020) \cite{phan2020epc} & - & 88.93 & 3.09$\times$ & \ding{55} & \ding{55} & \ding{55} \\
BATUDE (2022)\cite{batude2022aaai} & - & 89.41 & 2.52$\times$ & \ding{51} & \ding{55} & \ding{55} \\
ORTOS Tucker (2024)\cite{aghababaeiharandi2024unified} & 70.88 & 89.87 & 3.03$\times$ & \ding{55} & \ding{51} & \ding{55} \\
APNN (2025)\cite{liu2025accuracypreserving} & - & 89.04 & 3.22$\times$ & \ding{55} & \ding{55} & \ding{55} \\
\rowcolor{green!10}
\textbf{CompressNAS} & \textbf{71.00} & \textbf{90.55} & \textbf{2.52$\times$} & \ding{51} & \ding{51} & \ding{51} \\
\rowcolor{green!10}
\textbf{CompressNAS} & \textbf{70.78} & \textbf{90.19} & \textbf{3.00$\times$} & \ding{51} & \ding{51} & \ding{51} \\
\hline
\end{tabular}
\end{table*}

\begin{table}[h!]
\centering
\small 
\caption{Comparison of lightweight models on ImageNet-1K \cite{deng2009imagenet} with $<4$M parameters.}
\label{tab:light_models_sota}
\rowcolors{2}{gray!10}{white}
\begin{tabular}{l c c}
\hline
\rowcolor{gray!10}
\textbf{Model} & \textbf{Params (M)} & \textbf{Top-1 Acc. (\%)} \\
\hline

MobileNetV1-1.00 \cite{howard2017mobilenets} & 4.20 & 70.6 \\
\rowcolor{orange!10}
MobileNetV3-large-0.75 \cite{howard2019mobilenetv3} & 4.00 & 73.3 \\
FasterNet-T0 \cite{wang2023fasternet} & 3.90 & 71.9 \\
MobileNetV2-1.00 \cite{howard2017mobilenets}& 3.40 & 71.8  \\
\rowcolor{green!10}
\textbf{STResNet-tiny} & 3.99 & 71.6 \\

\hline 
MobileNetV3-small-1.0 \cite{howard2019mobilenetv3} & 2.90 & 67.4 \\

MobileNetV1-0.75 \cite{howard2017mobilenets} & 2.60 & 68.4 \\
MobileNetV3-small-0.75 \cite{howard2019mobilenetv3} & 2.40 & 65.4 \\
ShuffleNetV2 1.0$\times$ \cite{ma2018shufflenetv2} & 2.30 & 69.4 \\
MNASNet0.5 \cite{tan2019mnasnetplatformawareneuralarchitecture} & 2.15 & 67.8 \\

MobileOne-S0 \cite{wang2023fasternet} & 2.10 & 71.4 \\
MobileNetV2-0.5 \cite{sandler2018mobilenetv2} & 2.00 & 65.4 \\

MobileNetV2-0.35 \cite{sandler2018mobilenetv2} & 1.70 & 60.3 \\
\rowcolor{green!10}
\textbf{STResNet-micro} & 1.50 & 66.7 \\

\hline

ShuffleNetV2 0.5$\times$ \cite{ma2018shufflenetv2} & 1.40 & 61.0 \\
MobileNetV1-0.5 \cite{howard2017mobilenets} & 1.30 & 63.7 \\
SqueezeNet 1.0 \cite{iandola2016squeezenet} & 1.25 & 57.5 \\
SqueezeNet 1.1 \cite{iandola2016squeezenet} & 1.24 & 58.2 \\
\rowcolor{orange!10}
ShuffleNetV1 1.0$\times$ \cite{zhang2018shufflenet} & 1.00 & 65.9 \\
\rowcolor{green!10}
\textbf{STResNet-nano} & 0.95 & 58.8 \\

\hline 

\rowcolor{orange!10}
ShuffleNetV1 0.5$\times$ \cite{zhang2018shufflenet} & 0.50 & 52.5 \\
MobileNetV1-0.25 \cite{howard2017mobilenets} & 0.47 & 41.5 \\
\rowcolor{green!10}
\textbf{STResNet-pico} & 0.62 & 48.8 \\
\hline
\end{tabular}
\end{table}

\begin{figure}
    \centering
    \includegraphics[width=0.9\linewidth]{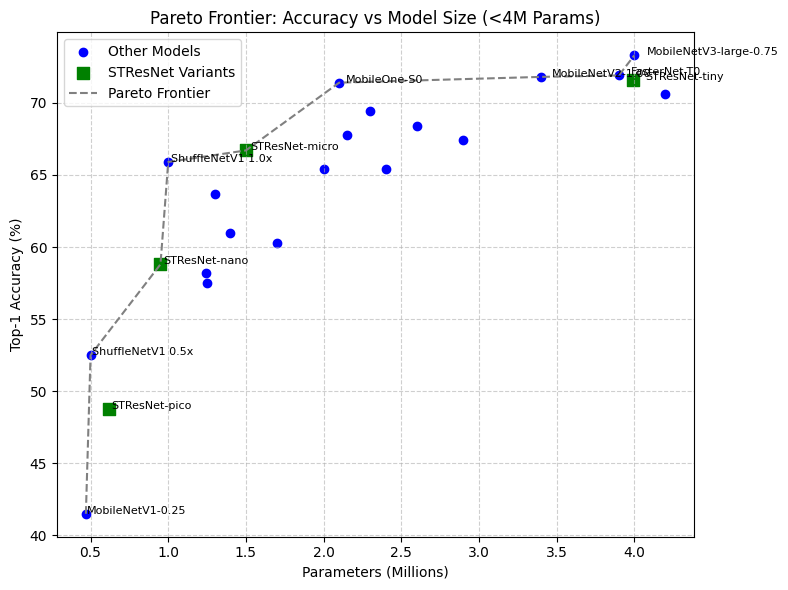}
    \caption{Pareto tiny Models compared to STResNet family of models}
    \label{fig:pareto_models}
\end{figure}

\subsection{Optimization Efficiency}
A key advantage of \emph{CompressNAS} is substantial reduction in overall optimization time, 
enabled by its low-cost accuracy and flash estimators plus the fast ILP search \cite{PuLP}. 
The accuracy and flash lookup tables can be generated on an AMD EPYC 7542 32-Core processor 
in under 30 minutes. The subsequent ILP search under any given constraint produces 
a solution within seconds. Consequently, the end-to-end optimization time for a model 
consists of one full training cycle plus approximately 30 minutes.  

For repeated optimization of the same model with different objectives, 
the additional computation cost is negligible, as the lookup tables are already available. 
In contrast, traditional tucker-based rank estimation requires comparable training effort 
but yields significantly larger accuracy degradation. Relative to Neural Architecture Search (NAS) methods, CompressNAS further improves efficiency:
state-of-the-art NAS approaches typically require 3--4 full training cycles to achieve
results comparable to CompressNAS, with a search complexity of approximately
\(\mathcal{O}(n \cdot E)\), where \(E\) denotes the number of fine-tuning epochs
per candidate. This stark difference arises from the methodology: 
NAS-based strategies rely on partial fine-tuning to approximate the impact of decomposition, 
whereas CompressNAS leverages zero-cost estimators in combination with the MicroNAS strategy 
to directly guide rank selection.

\subsection{State-of-the-Art Comparison}

Table \ref{tab:quark_sota_comparison} compares state-of-the-art compression methods on ResNet-18, focusing on low-rank decomposition approaches (Vanilla Tucker, MUSCO, Stable EPC, ORTOS) and our proposed method, CompressNAS. Metrics reported are Top-1 and Top-5 accuracy on ImageNet \cite{deng2009imagenet}, along with compression ratio (parameter reduction).

\emph{CompressNAS} achieves the Top-1 (71.84\%) and Top-5 (90.54\%) accuracies with a competitive 2.52× compression, demonstrating that its global search-based rank selection outperforms prior heuristic, decomposition-based, and recent fine-grain search methods \cite{aghababaeiharandi2024unified}. Other methods, such as BATUDE and Stable EPC, achieve similar compression with either comparable or lower accuracy, highlighting \emph{CompressNAS}’s ability to balance efficiency and performance. ORTOS \cite{aghababaeiharandi2024unified} and Accuracy preserving Neural Network Compression (APNN) \cite{liu2025accuracypreserving} achieve approximately similar performance given that our experiment is for lower compression (all state of the art methods use different level of compression so it is not practical to compare with all results with same compression). However, APNN proposed a joint optimization during model training which means the optimziation needs to run again for a different budget. The concept is same for budget aware compression technique \cite{batude2022aaai} where fine grain rank selection is performed at the time of model training. Our optimization method runs once and the model can be generated for different hardware and constraint for different dataset.   

In addition to accuracy preservation, \emph{CompressNAS} supports budget-aware optimization \cite{batude2022aaai} and fine-grain rank selection \cite{unified2024arxiv} while introducing a configurable optimization framework: the search is executed once and can be reused to generate models at different compression levels, enhancing practical deployment flexibility as shown in \ref{tab:quark_sota_comparison}. We have not compared the performance of our model with \cite{liu2025accuracypreserving}.

\subsection{STResNet}

We present a family of models generated using \emph{CompressNAS} called \emph{STResNet} (being the variant of ResNet). Table \ref{tab:light_models_sota} presents a comparison of our STResNet family of models with existing lightweight CNNs on ImageNet-1K \cite{deng2009imagenet} under the constraint of fewer than 4M parameters. We designed four variations of STResNet (\textit{Tiny, Micro, Nano and Pico} according to model size, and across most of the scales, STResNet consistently achieves a strong trade-off between accuracy and parameter efficiency. In particular, STResNet-tiny attains 71.6\% Top-1 accuracy with fewer than 4M parameters, closely matching MobileNetV2 (71.8\%) but lower performance compared to MobileNetV3-large-0.75.

At smaller scales, STResNet-micro demonstrates state-of-the-art performance compared to other models in the same parameter range. In the compact segment, SqueezeNet 1.0 slightly outperforms STResNet-nano, whereas in the ultra-compact regime, STResNet-pico surpasses MobileNetV1-0.25 by over 7\% Top-1 accuracy with only 0.62M parameters, although it falls slightly behind ShuffleNetV1 0.5×. These results indicate that STResNet variants push the Pareto frontier of accuracy versus model size (see Fig. \ref{fig:pareto_models}), making them well-suited for resource-constrained deployments.

Another important consideration is the quantization-friendliness of compact models such as MobileNet \cite{howard2017mobilenets, sandler2018mobilenetv2, howard2019mobilenetv3}, SqueezeNet \cite{iandola2016squeezenet}, EfficientNet \cite{tan2019efficientnet}, and ShuffleNet \cite{ma2018shufflenetv2, zhang2018shufflenet}. Many of these models experience substantial accuracy drops under ultra-low-bit quantization (less than 8-bit) as reported in prior work \cite{li2021brecq, park2020profit}. SqueezeNet and EfficientNet are often excluded from such studies due to inherent quantization challenges, typically requiring quantization-aware training (QAT) to recover accuracy. These difficulties arise from specialized layers designed for efficiency.

In contrast, STResNet is a decomposed version of ResNet \cite{he2016deep}, whose simple and regular architecture is highly amenable to quantization. STResNet is specifically designed for deployment on small devices such as microcontroller units (MCUs) and neural processing units (NPUs), where low-bit quantization is a primary constraint. This makes STResNet particularly competitive in scenarios demanding both compactness and quantization efficiency.

\section{Conclusion}
\label{sec:conclusion}

In this work, we introduced \emph{CompressNAS}, a microNAS-based decomposition framework leveraging zero-cost estimators for efficient rank selection and model compression. Across classification (ResNet-18, ResNet-34, MobileNetV2) and detection tasks (YOLOv5), \emph{CompressNAS} achieves high compression ratios while retaining accuracy, either outperforming or at par with other tucker based decomposition methods presented in literature. On ResNet-18, it attains among the highest Top-1 and Top-5 accuracies compared to state-of-the-art approaches, demonstrating the effectiveness of global search-based rank selection. \emph{CompressNAS} also supports budget-aware and fine-grain rank optimization, with a configurable framework allowing a single search to generate models at different compression levels \cite{aghababaeiharandi2024unified}, significantly reducing overhead. These results establish \emph{CompressNAS} as a practical, scalable, and flexible solution for deploying models on resource-constrained devices, effectively balancing aggressive compression with real-world applicability. At the end, we present \emph{STResNet} a family of tiny models demonstrating competitive performance under 4M parameter budget.  

\section*{Acknowledgments}

We acknowledge our colleagues at STMicroelectronics: Ehsan Saboori and Ravish Kumar; Olivier Mastropietro and Alexander Hoffman from Deeplite  for their prior contributions to Tucker decomposition research, which inspired aspects of this work.
\bibliographystyle{unsrt}
\bibliography{references}

\null
\thispagestyle{empty}
\newpage

\section{Appendix}
\subsection{STResNet Family Architecture}
\begin{table}[h!]
\centering
\small
\caption{STResNet-Pico Architecture }
\label{tab:stresnet_pico_compact}
\begin{tabular}{l c c}
\hline
\textbf{Block / Layer} & \textbf{In $\to$ Out} & \textbf{Conv Layers (Inner Channels)} \\
\hline
\textbf{Stem} & 3 $\to$ 64 & 1x1/3→3, 7x7/3→8, 1x1/8→32 \\ & & \textbf{proj.1x1/32→64} \\
\hline
\textbf{Layer1} & 64 $\to$ 64 & Block1: 1x1/64→24, 3x3/24, 1x1/24→64; 1x1/64→16, 3x3/16, 1x1/16→64 \\
& & Block2: 1x1/64→24, 3x3/24, 1x1/24→64; 1x1/64→8, 3x3/8, 1x1/8→64 \\
\hline
\textbf{Layer2} & 64 $\to$ 128 & Block1: 1x1/64→24, 3x3/24 s2, 1x1/24→128; 1x1/128→8, 3x3/8, 1x1/8→128 \\
& & Block2: 1x1/128→8, 3x3/8, 1x1/8→128; 1x1/128→8, 3x3/8, 1x1/8→128 \\
\hline
\textbf{Layer3} & 128 $\to$ 256 & Block1: 1x1/128→8, 3x3/8 s2, 1x1/8→256; 1x1/256→8, 3x3/8, 1x1/8→256 \\
& & Block2: 1x1/256→8, 3x3/8, 1x1/8→256; 1x1/256→8, 3x3/8, 1x1/8→256 \\
\hline
\textbf{Layer4} & 256 $\to$ 512 & Block1: 1x1/256→8, 3x3/8 s2, 1x1/8→512; 1x1/512→8, 3x3/8, 1x1/8→512 \\
& & Block2: 1x1/512→8, 3x3/8, 1x1/8→512; 1x1/512→8, 3x3/8, 1x1/8→512 \\
\hline
\end{tabular}
\end{table}

\begin{table}[h!]
\centering
\small
\caption{STResNet-Tiny Architecture}
\label{tab:stresnet_tiny_compact}
\begin{tabular}{l c c}
\hline
\textbf{Block / Layer} & \textbf{In $\to$ Out} & \textbf{Conv Layers (Inner Channels)} \\
\hline
\textbf{Stem} & 3 $\to$ 64 & 1x1/3→3, 7x7 s2/3→16, 1x1/16→32 \\ & & \textbf{proj.1x1/32→64} \\ 
\hline
\textbf{Layer1} & 64 $\to$ 64 & Block1: 3x3/64→64, 3x3/64→64 \\
& & Block2: 3x3/64→64, 3x3/64→64 \\
\hline
\textbf{Layer2} & 64 $\to$ 128 & Block1: 3x3 s2/64→128; 1x1/128→96, 3x3/96, 1x1/96→128 \\
& & Block2: 3x3/128→128; 1x1/128→80, 3x3/80, 1x1/80→128 \\
\hline
\textbf{Layer3} & 128 $\to$ 256 & Block1: 3x3 s2/128→256; 1x1/256→192, 3x3/192, 1x1/192→256 \\
& & Block2: 3x3/256→256; 1x1/256→96, 3x3/96, 1x1/96→256 \\
\hline
\textbf{Layer4} & 256 $\to$ 512 & Block1: 1x1/256→208, 3x3 s2/208→208, 1x1/208→512; \\ & &  1x1/512→88, 3x3/88, 1x1/88→512 \\
& & Block2: 1x1/512→192, 3x3/192, 1x1/192→512; \\ & & 1x1/512→112, 3x3/112, 1x1/112→512 \\
\hline
\end{tabular}
\end{table}

\begin{table}[h!]
\centering
\small
\caption{STResNet-Micro Architecture }
\label{tab:stresnet_micro_compact}
\begin{tabular}{l c c}
\hline
\textbf{Block / Layer} & \textbf{In $\to$ Out} & \textbf{Conv Layers (Inner Channels)} \\
\hline
\textbf{Stem} & 3 $\to$ 64 & 1x1/3→3, 7x7/3→8, 1x1/8→32 \\ & & \textbf{proj.1x1/32→64} \\
\hline
\textbf{Layer1} & 64 $\to$ 64 & Block1: 1x1/64→64, 3x3/64, 1x1/64→64; 1x1/64→64, 3x3/64, 1x1/64→64 \\
& & Block2: 1x1/64→64, 3x3/64, 1x1/64→64; 1x1/64→64, 3x3/64, 1x1/64→64 \\
\hline
\textbf{Layer2} & 64 $\to$ 128 & Block1: 1x1/64→40, 3x3/40 s2, 1x1/40→128; 1x1/128→32, 3x3/32, 1x1/32→128 \\
& & Block2: 1x1/128→88, 3x3/88, 1x1/88→128; 1x1/128→32, 3x3/32, 1x1/32→128 \\
\hline
\textbf{Layer3} & 128 $\to$ 256 & Block1: 1x1/128→88, 3x3/88 s2, 1x1/88→256; 1x1/256→72, 3x3/72, 1x1/72→256 \\
& & Block2: 1x1/256→80, 3x3/80, 1x1/80→256; 1x1/256→32, 3x3/32, 1x1/32→256 \\
\hline
\textbf{Layer4} & 256 $\to$ 512 & Block1: 1x1/256→80, 3x3/80 s2, 1x1/80→512; 1x1/512→8, 3x3/8, 1x1/8→512 \\
& & Block2: 1x1/512→72, 3x3/72, 1x1/72→512; 1x1/512→64, 3x3/64, 1x1/64→512 \\
\hline
\end{tabular}
\end{table}

\end{document}